\documentclass{egpubl}
\usepackage{pg2024}

\SpecialIssuePaper         

\usepackage[T1]{fontenc}
\usepackage{dfadobe}  

\usepackage{cite}  
\BibtexOrBiblatex
\electronicVersion
\PrintedOrElectronic
\ifpdf \usepackage[pdftex]{graphicx} \pdfcompresslevel=9
\else \usepackage[dvips]{graphicx} \fi

\usepackage{egweblnk} 

\title{GGAvatar: Geometric Adjustment of Gaussian Head Avatar}

\author[Xinyang Li, Jiaxin Wang, Yixin Xuan, Gongxin Yao, Yu Pan]
{\parbox{\textwidth}{\centering Xinyang Li\thanks{l$\_$xyang@zju.edu.cn}$^{1}$\orcid{0009-0000-5818-7822}
        , Jiaxin Wang$^{2}$\orcid{0009-0005-1632-5197}
        , Yixin Xuan$^{1}$
        , Gongxin Yao$^{1}$
        , Yu Pan$^{1}$
        }
        \\
{\parbox{\textwidth}{\centering $^1$Zhejiang University, Hangzhou, China\\
         $^2$Hangzhou Dianzi University, Hangzhou, China
       }
}
}

\usepackage{amsfonts}
\usepackage{amsmath}
\usepackage{bm}
\usepackage{amssymb}  
\usepackage{booktabs} 
\usepackage{graphicx} 
\usepackage[table]{xcolor} 
\usepackage{colortbl} 
\begin{document}


\maketitle
\begin{abstract}
  We propose GGAvatar, a novel 3D avatar representation designed to robustly model dynamic head avatars with complex identities and deformations.  GGAvatar employs a coarse-to-fine structure, featuring two core modules: Neutral Gaussian Initialization Module and Geometry Morph Adjuster. Neutral Gaussian Initialization Module pairs Gaussian primitives with deformable triangular meshes, employing an adaptive density control strategy to model the geometric structure of the target subject with neutral expressions. Geometry Morph Adjuster introduces deformation bases for each Gaussian in global space, creating fine-grained low-dimensional representations of deformation behaviors to address the limitations of the Linear Blend Skinning formula effectively. Extensive experiments show that GGAvatar can produce high-fidelity renderings, outperforming state-of-the-art methods in both visual quality and quantitative metrics.


\printccsdesc   
\end{abstract}  
\section{Introduction}
Creating high-fidelity digital avatars with real-time interaction accessible to everyone is a crucial building block for the metaverse and various applications, such as immersive telepresence and augmented or virtual reality. Although achieving photorealistic, deformable digital avatars has been a research focus in computer vision and graphics for decades, generalizing these avatars to unseen poses or expressions with low cost remains an ongoing challenge.

Given a personalized video, most prior works have primarily employed 3D Morphable Model (3DMM) techniques \cite{li2017learning, paysan20093d} or have leveraged neural implicit representations \cite{mildenhall2021nerf, park2019deepsdf} to develop animable 3D head avatars. The former method \cite{grassal2022neural, khakhulin2022realistic, kim2018deep} employs a fixed topological structure linked to a standard rasterization pipeline, enabling it to generalize to unseen deformations. However, this fixed topology lacks the structural flexibility for accessories such as hats and glasses, thereby restricting its ability to model individuals with complex identities accurately. Works based on neural implicit representations \cite{bergman2022generative, hong2022headnerf, zheng2022avatar} leverage the sampling of multiple points along camera rays to capture accessories. However, these approaches significantly reduce their training and rendering efficiency. 

3D Gaussians Splatting (3D-GS) \cite{kerbl20233d} recently has proven more efficient than Neural Radiance Fields (NeRF) \cite{mildenhall2021nerf} in new perspective synthesis and 3D reconstruction. It represents 3D scenes explicitly by introducing discrete 3D Gaussian primitives and adapts seamlessly to the rasterization pipeline, thereby providing real-time rendering performance. Motivated by this progress, there has been significant research \cite{chen2023monogaussianavatar, zhao2024psavatar, qian2023gaussianavatars, xiang2023flashavatar} aimed at extending these innovations to the development of 3D digital avatars. Despite substantial advancements, two major challenges persist with current Gaussian-based methods: 1) Given their discrete nature, how can one properly initialize the geometry to accelerate the convergence of training? Flashavatar \cite{xiang2023flashavatar} addresses this by transforming the 3D head into UV space for sampling, achieving high-fidelity digital heads at 300FPS. However, this sampling strategy, heavily reliant on the FLAME model \cite{li2017learning}, struggles to depict features such as long hair or shoulders accurately. 2) How can a deformation strategy be designed to generalize 3D avatars to unseen poses and expressions effectively? An intuitive approach employs parameterized 3DMM of human heads as drivers. These models provide several orthogonal bases within their parameter space, enabling the manipulation of attributes such as identity, pose, and expression in the 3D model via linear blend skinning (LBS). Some approaches \cite{qian2023gaussianavatars, shao2024splattingavatar} leverage this strategy by anchoring the Gaussians to a 3D mesh, enabling localized learning of intrinsic features, and dynamically adjusting their global properties, such as position, scaling, and rotation, based on alterations in the 3D model's topology. However, this strategy relies on the geometric consistency of multiple views. It cannot be extended to exaggerated expressions and poses through direct linear transformations due to the limitations of the LBS formula, thereby restricting the animation capabilities of 3D avatars.

\begin{figure}
    \centering
    \includegraphics[width=1\linewidth]{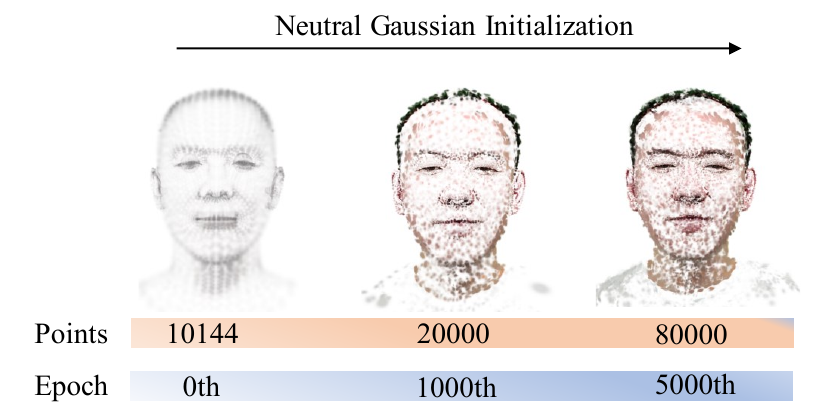}
    \caption{The initialization process for a neutral expression (e.g., ID1) uses a densification strategy to add Gaussian primitives to non-head regions, accelerating training convergence. This method shows that even without corresponding neutral expression images, we can reconstruct the neutral Gaussian geometry using the binding Gaussian strategy.}
    \label{fig:1}
\end{figure}

In this paper, we introduce a novel 3D avatar representation, called GGAvatar, designed to model dynamic head avatars with complex identities and deformations robustly. GGAvatar consists of two core modules: Neutral Gaussian Initialization module (defined in this article as global Gaussians without facial parameter inputs) and Geometry Morph Adjuster. For the Neutral Gaussian Initialization module, We initialize a 3D avatar featuring a neutral expression and appearance and construct a motion offset field to integrate high-frequency dynamic details and head movements. To be concrete, we bind Gaussians to the FLAME mesh like the GaussianAvatar \cite{qian2023gaussianavatars} approach and employ the densification strategy of 3D-GS \cite{kerbl20233d} to initialize the 3D avatar with neutral expressions. This initialization technique swiftly enriches areas of the avatar that lack detail with Gaussian primitives, as shown in Fig~\ref{fig:1}. For deformation purposes, we harness bound Gaussians that track the coarse movements of meshes to ensure stability throughout the animation process. However, methods based on LBS formula deformation cannot capture the motion of fine non-surface regions, such as wrinkles and hair. To address the limitations, we propose the Geometry Morph Adjuster. The Geometry Morph Adjuster employS a parameterized multi-resolution tri-plane to store the spatial information of neutral Gaussians, connecting it to a finely-tuned Multi-Layer Perceptron (MLP) that learns the deformation bases for each Gaussian. This setup facilitates the creation of a low-dimensional representation of deformation behaviors for each Gaussian. Integrating this with pre-retrieved facial parameters allows us to predict further positional adjustments and covariance shifts, significantly boosting the avatar's realism and expressiveness. By adopting this strategy, we mitigate the unpredictability of directly forecasting deformations and surpass existing works in terms of rendering novel views and reenactments from a driving video.

The contributions of our method can be summarized as:
\begin{itemize}
    \item We propose the GGAvatar, a novel representation that employs a neutral Gaussian initialization strategy and a coarse-to-fine architecture to model ultra-high-fidelity human head avatars.
    \item To capture high-frequency dynamic details, we utilize explicit parameterized tri-plane structures connected to a tiny MLP that learns deformation bases, which accurately model extremely complex and exaggerated facial deformation.
    \item Extensive testing on public datasets has demonstrated that our method outperforms contemporary alternatives in both visual quality and quantitative metrics.
\end{itemize}

\section{Related Work}
\subsection{Head Avatar Reconstruction with Implicit Models}
NeRF encapsulates the radiance field of a scene within a neural network, enabling photorealistic renderings of novel views through volumetric rendering techniques. \cite{gafni2021dynamic, gao2021dynamic, wang2021learning, wang2021prior} directly manipulates NeRF using facial model parameters such as expressions and poses to create an animable head avatar. However, these methods struggle to disentangle poses and expressions effectively, and they also face challenges in generalizing to novel poses and expressions. An alternative strategy \cite{kocabas2023hugs, ye2023animatable} employs the "canonical + deformation" framework to construct a head model in a standard space and generate dynamics via deformation fields. 
IMAvatar \cite{zheng2022avatar} employs a signed distance function to represent the implicit head model and uses learnable expression blend shapes to depict the deformation field. Additionally, various methods \cite{athar2022rignerf, bergman2022generative, athar2023flame, gafni2021dynamic, liu2022semantic} leverage advanced techniques to enhance training speed and rendering quality, such as TriPlane \cite{zhao2023havatar}, KPlane \cite{fridovich2023k}, and deformable multi-layer meshes \cite{duan2023bakedavatar}. Nonetheless, these techniques still grapple with challenges related to efficiency in training and rendering.

\subsection{Head Avatar Reconstruction with Explicit Models}
3DMM \cite{Blanz_Vetter_1999} initially projects the 3D head shape into several low-dimensional Principal Component Analysis (PCA) spaces.
Subsequently, numerous works have adopted 3DMM and its variants \cite{cao2013facewarehouse, feng2021learning, paysan2009face, gerig2018morphable} to create head avatars. \cite{feng2021learning, thies2019deferred, thies2020neural} employ feed-forward networks to predict vertex offsets and textures, enabling inference of unseen facial expressions. ROME \cite{khakhulin2022realistic} encodes local photometric and geometric details to improve the quality of rendered images. 3DMM-based methods provide stable deformations but fail to model accessories such as eyeglasses. PointAvatar \cite{zheng2023pointavatar} introduces a point-based geometric representation using differentiable point rendering, which overcomes the limitations of mesh-based models. However, it requires an excessive number of points and extensive training periods.
\begin{figure*}
    \centering
    \includegraphics[width=1\linewidth]{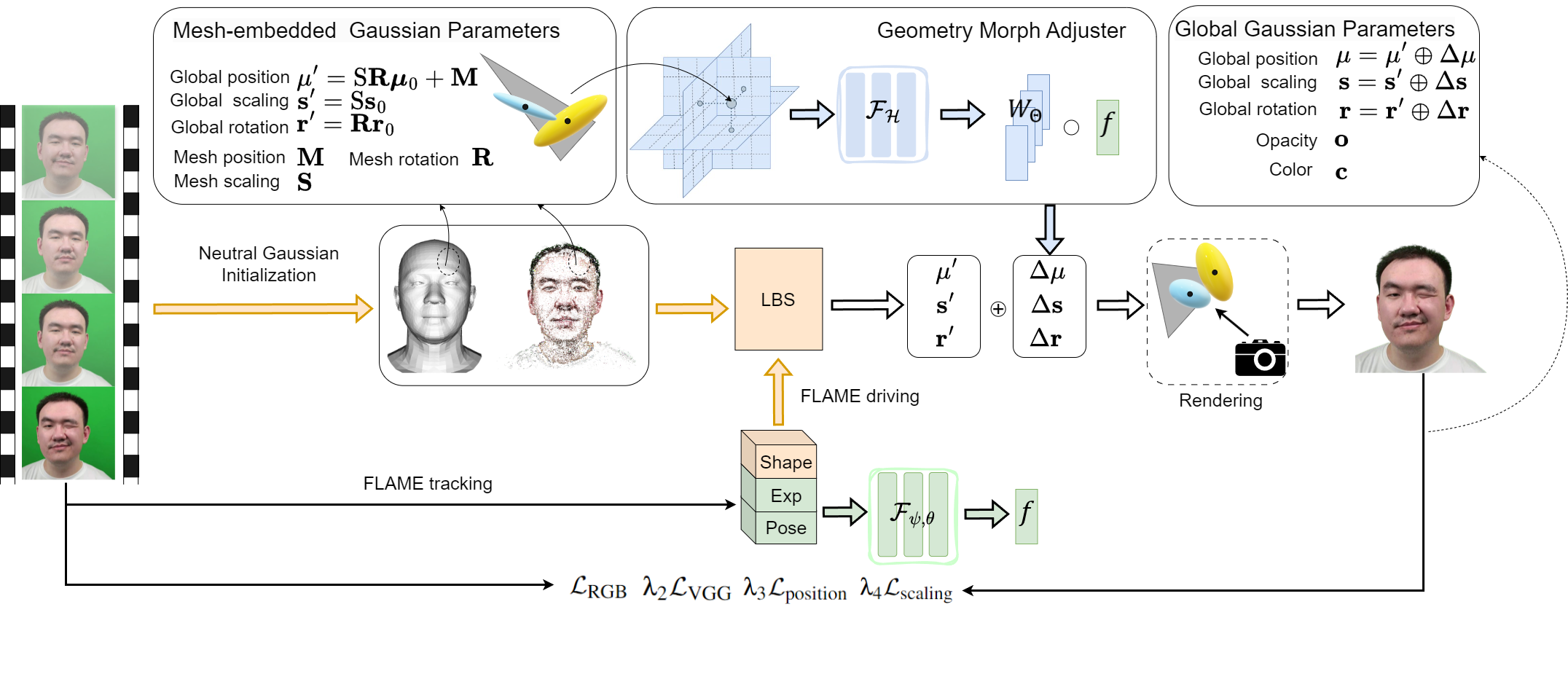}
    \caption{Overview of GGAvatar. A mesh-embedded Gaussian initialization strategy is proposed to model the geometry of neutral Gaussians. The neutral Gaussians are then coarsely deformed with FLAME mesh. To capture high-frequency dynamic details, we introduce the Geometry Morph Adjuster.  To further enhance the representation capability of the deformation bases, we generate a latent vector from expression and pose parameters using an MLP. The deformed Gaussians are then splatted to render the image with a given pose.}
    \label{fig:2}
\end{figure*}
Recently, 3D-GS \cite{kerbl20233d} has demonstrated exceptional performance, offering greater flexibility due to its anisotropic properties compared to point representations, and providing the efficiency of real-time rendering. Expanding upon PointAvatar, MonoGaussianAvatar \cite{chen2023monogaussianavatar} replaces Point Cloud with Gaussian primitives to improve the rendering speed and quality. PSAvatar \cite{zhao2024psavatar} further proposes a point-based morphable shape model to increase the flexibility of representation. Despite the impressive results achieved by these methods in rendering novel viewpoints and reenactments, the point-based initialization requiring redefinition of Gaussian features adds complexity to the training process. FLashAvatar \cite{xiang2023flashavatar} samples in UV space and attaches Gaussians to mesh with learnable offsets, which are represented as MLPs, enabling high-speed rendering of human head avatars. However, this initialization strategy distributes Gaussians across the FLAME model, limiting its ability to model features such as long hair and shoulders effectively. GaussianAvatars \cite{chen2023monogaussianavatar} associates each mesh triangle with a 3D Gaussian, incorporating densification and pruning strategies for accurate geometry representation, and uses binding inheritance to ensure seamless animation control via the parametric model. However, deformations relying on the LBS formula fail to animate intricate non-surface structures like hair and wrinkles. Therefore, building on the GaussianAvatar binding and the 3D-GS densification strategy \cite{kerbl20233d}, we introduce a parametric three-plane hash structure to capture more detailed deformation offsets, enabling 
high-fidelity animation of head avatars.
\section{Method}
Figure~\ref{fig:2} presents a schematic of GGAvatar. Given a monocular portrait video of a target subject, the objective is to reconstruct an animatable head avatar. The FLAME meshes \cite{li2017learning} feature vertices at varying positions but share the same topology. Consequently, we pair 3D Gaussian primitives with triangles of the mesh, employing a densification strategy \cite{kerbl20233d} to geometrically initialize the 3D head avatar and facilitate coarse Gaussian deformations (see section~\ref{sec:3.2}). 
Additionally, we learn an extra deformation basis for each Gaussian to achieve fined deformations and enhance high-frequency details (see section~\ref{sec:3.3}). 

%

\subsection{Preliminary}\label{sec:3.1}
\noindent 3D-GS utilizes discrete Gaussian primitives for the geometry representation of static scenes. A Gaussian primitive is defined by a 3D covariance matrix $\bm{\Sigma}$ centered at point (mean) $\mu$:
\begin{equation}\label{eq:1}
G(\mathbf{x}) = e^{-\frac{1}{2} (\mathbf{x} - \mu)^T \bm{\Sigma}^{-1} (\mathbf{x} - \mu)}.
\end{equation}
To ensure differentiable computation of the covariance matrix, \cite{kerbl20233d} defines a parametric ellipsoid with a scaling matrix $\textbf{S}$ and a rotation matrix $\textbf{R}$, constructing the covariance matrix by:
\begin{equation}\label{eq:2}
\bm{\Sigma}=\operatorname{\textbf{RSS}}^T\operatorname{\textbf{R}}^T.
\end{equation}
In particular, the scaling and rotation matrices are represented by a learnable scaling vector $\textbf{s} \in \mathbb{R}^3$ and a learnable quaternion $\textbf{r} \in \mathbb{R}^4$, respectively. 
Finally, Finally, the color of a pixel can be determined through the blending of overlapped Gaussians:
\begin{equation}\label{eq:3}
\mathbf{C}=\sum_{i \in N} \mathbf{c}_i \alpha_i \prod_{j=1}^{i-1}\left(1-\alpha_j\right),
\end{equation}
where $\mathbf{c}_i$ is the color of each point represented by the spherical harmonic function, and blending weight $ \alpha_i $ is computed by the 2D projection of the 3D Gaussian multiplied by a per-point opacity $o$.

\subsection{Neutral Gaussian Initialization}\label{sec:3.2}
Given a set of tracked FLAME triangular meshes with vertices $\{\mathbf{v}_0, \mathbf{v}_1, \mathbf{v}_2\}$ and edges defined as $\{\mathbf{a}_{10} = \mathbf{v}_1 - \mathbf{v}_0$, $\mathbf{a}_{20} = \mathbf{v}_2 - \mathbf{v}_0$, and $\mathbf{a}_{21} = \mathbf{v}_2 - \mathbf{v}_1\}$, we initialize each Gaussian at the centroid of these meshes $\mathbf{M}$. Specifically, we establish that $\mathbf{M}$ is the coordinate origin for each Gaussian in its local space. Then, We establish that the rotation matrix $\mathbf{R}$ represents the orientation of the triangle in global space, and the scaling $\mathrm{S}$ reflects the average size of the triangle as follows:
\begin{align}\label{eq:4}
\mathbf{R} &= \left[\mathbf{n}_0 ; \mathbf{n}_1 ; \mathbf{n}_2\right],\\
\mathrm{S} &= \left(\left\|\mathbf{a}_{20}\right| + \left|\mathbf{n}_2 \cdot \mathbf{a}_{21}\right\|\right)/2,
\end{align}
where $\mathrm{S}$ is represented as the average length of one edge vector and its perpendicular distance, and $\left[\mathbf{n}_0; \mathbf{n}_1; \mathbf{n}_2\right]$ respectively represent the unit vector along $\mathbf{a}_{10}$, the unit normal vector of the mesh, and the cross product of $\mathbf{n}_0$ and $\mathbf{n}_1$.

According to Sec. \ref{sec:3.1}, we parameterize all its properties in local space, defined as $G=\{\mu_0, \mathbf{r}_0, \mathbf{s}_0, o, \mathbf{c}\}$. In the rough rendering stage, we convert these properties into the global space by:
\begin{align}
\mathbf{r}^{\prime} & =\mathbf{R} \mathbf{r}_0, \\
\mu^{\prime} & =\mathrm{S} \mathbf{R} \boldsymbol{\mu}_0+\mathbf{M}, \\
\mathbf{s}^{\prime} & =\mathrm{S} \mathbf{s}_0 .
\end{align}
To better capture the geometric shape of human head avatars, we employ an adaptive density control strategy \cite{kerbl20233d}, adding and removing splats based on the gradient of the view-space position and the opacity of each Gaussian. Then, we bind new Gaussians to the old ones during the optimization process to restore fidelity in local areas.

The motion of head avatars can be divided into rigid movements related to head pose and non-rigid transformations associated with facial expressions. Head pose, including rotation and translation, can be manipulated through camera parameters. Based on this observation, during the initialization phase, we input zero expression parameters to control the local learning of Gaussians. By computing images of various expressions from the training data and using backpropagation to adjust the properties of the neutral Gaussians, we can model the geometric structure of neutral Gaussians even without corresponding ground truth inputs for zero expressions.
\subsection{Geometry Morph Adjuster}\label{sec:3.3}
Gaussians moving with the meshes is sufficient to capture coarse head movements; however, constrained by the direct linear representation of LBS, they are insufficient for capturing all the fine and intricate dynamic textures, such as hair and wrinkles. To enhance representational capacity while minimizing resource usage, we utilize a multi-resolution tri-plane to store high-frequency spatial information surrounding the head, denoted as $\mathcal{H}^3$. Specifically, we initially transfer the neutral Gaussians to global space and denote their positions as $\mathbf{x}$ with the tri-plane outputting $\mathcal{H}^3(\mathbf{x})$. Inspired by the dynamic representations of PointAvatar \cite{zheng2023pointavatar}, we learn an additional basis $\mathcal{W}_{\Theta}$ for each neutral Gaussian, aimed at achieving precise deformations for each Gaussian given conditional parameter inputs as follows:
\begin{equation}
    \mathcal{W}_{\Theta}=\mathcal{F}_\mathcal{H}(\mathcal{H}^3(\mathbf{x})),
\end{equation}
where $\mathcal{F}_\mathcal{H}$ is a basis prediction network represented as MLPs.

To reduce computational costs and enhance the accuracy of deformations, we employ another MLP $\mathcal{F}_{\psi, \theta}$ to reduce the dimensionality of the expression $\psi$ and pose $\theta$ parameters, representing them as a latent vector $\mathbf{\textit{f}}$:
\begin{equation}
    \mathbf{\textit{f}}=\mathcal{F}_{\psi, \theta}(\psi, \theta).
\end{equation}

Additional deformations of the Gaussian can be represented as the matmul product of the base and the latent vector:
\begin{equation}
    \Delta\mu, \Delta \mathbf{r}, \Delta \mathbf{s}= \mathcal{W}_{\Theta} \cdot \mathbf{\textit{f}}.
\end{equation}

The final refined spatial features can be calculated as:
\begin{equation}
\mu, \mathbf{r}, \mathbf{s}=\left(\mu^{\prime}  \oplus \Delta \mu, \mathbf{r}^{\prime}  \oplus \Delta \mathbf{r}, \mathbf{s}^{\prime} \oplus \Delta \mathbf{s}\right).
\end{equation}

It is important to note that we address the limitations of mesh motion by transforming Gaussians into global space and learning a set of additional offsets to deform each Gaussian, as detailed in Section 4 dynamically.
\subsection{Training Objectives}\label{sec:3.4}
For the loss function, we utilize L1 loss $\mathcal{L}_{1}$ to monitor the pixel-wise difference between the ground truth and the rendered images. Additionally, we incorporate a D-SSIM term $\mathcal{L}_{\text{D-SSIM}}$ to further ensure structural similarity between the two images:
\begin{equation}
\mathcal{L}_{\text{RGB}}=(1-\lambda_1) \mathcal{L}_{1} + \lambda_1 \mathcal{L}_{\text{D-SSIM}},
\end{equation}
where $\lambda_1$ is taken as 0.2, thanks to the powerful rendering technique \cite{kerbl20233d} that can process all the images during every training step, allowing us to apply perceptual losses $\mathcal{L}_{\text {VGG}}$ \cite{johnson2016perceptual} on the whole image. 

Besides, our rendering quality partially relies on the proper alignment between Gaussian splats and triangles; otherwise, unnatural jitter and artifacts may occur during video synthesis. To address this issue, To address this issue, we regularize the local position and scaling of each Gaussian to ensure they remain within reasonable limits:
\begin{align}
\mathcal{L}_{\text {position }}&=\left\|\max \left(\mu-\epsilon_{\text {position }, 
 0}\right)\right\|_2,\\
 \mathcal{L}_{\text {scaling }}&=\left\|\max \left(\mathbf{s}-\epsilon_{\text {scaling}, 
 0}\right)\right\|_2,
\end{align}
where $\epsilon_{\text {position}} = 1$ and $\epsilon_{\text {scaling}} = 0.6$ are the thresholds for the maximum allowable position and scaling, respectively. When below these thresholds, we disable their corresponding loss terms.

Our final loss function is thus:
\begin{equation}
\mathcal{L}=\mathcal{L}_{\text{RGB}} + \lambda_2 \mathcal{L}_{\text {VGG}}+\lambda_3 \mathcal{L}_{\text {position }} + \lambda_4 \mathcal{L}_{\text {scaling }},
\end{equation}
where $\lambda_2 = 0.02$, $\lambda_3 = 0.01$, and $\lambda_4 = 1$.
\begin{figure*}
    \centering
    \includegraphics[width=1\linewidth]{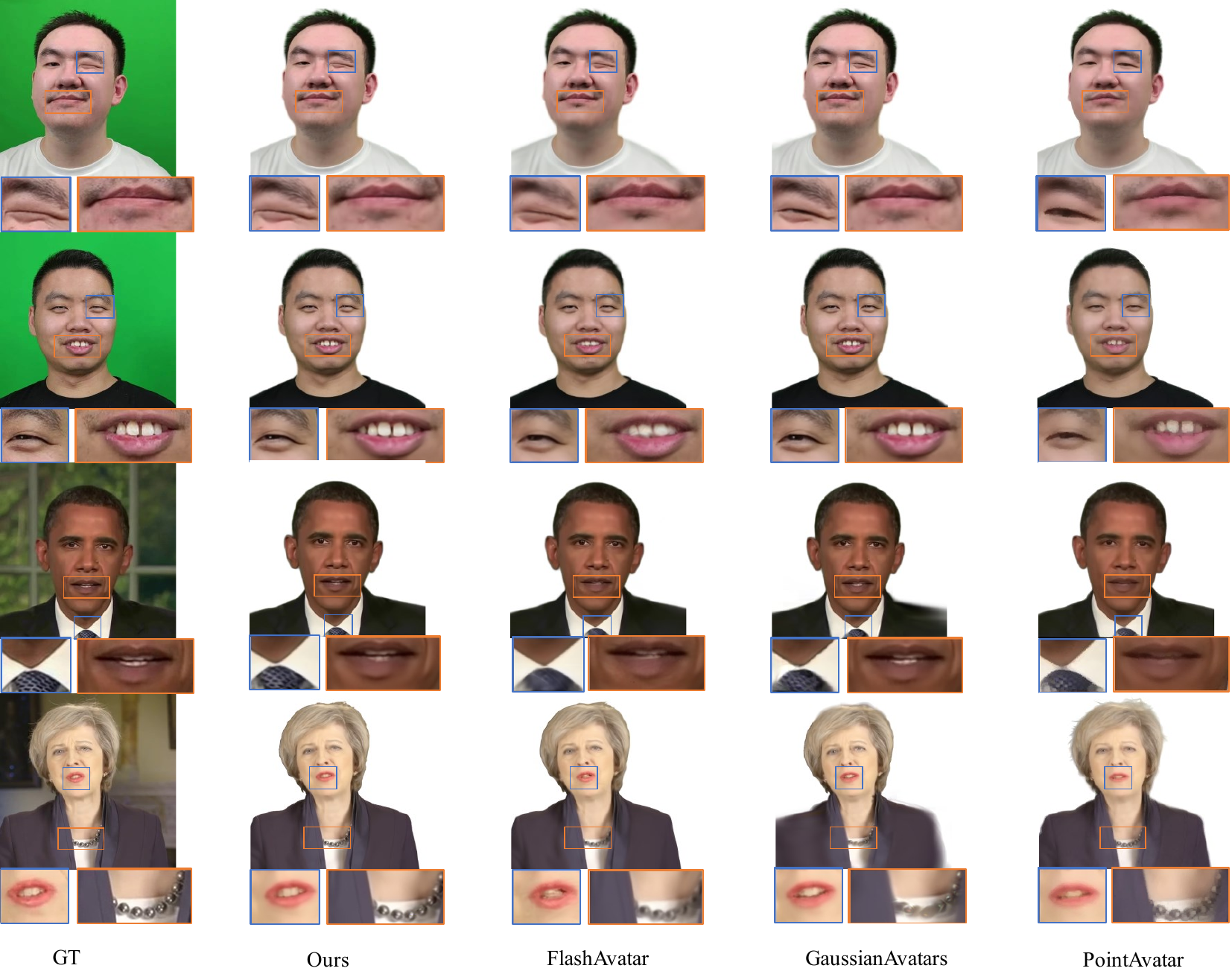}
    \caption{Qualitative Comparisons with State-of-the-Art Methods. From top to bottom is ID1, ID2, ID3, ID4. GGAvatar generates more realistic face reconstructions, especially in
capturing high-frequency dynamic details and reconstructing extreme expressions.
}
    \label{fig:3}
\end{figure*}
\subsection{Implementation Details}\label{sec:3.5}
We implement our network with Pytorch and use Adam for parameter optimization. We use the analysis-by-synthesis-based face tracker from MICA \cite{zielonka2022towards} for FLAME tracking. We conducted 120,000 training iterations for all target subjects and set all experiments to be performed on a single NVIDIA GTX 4090.

We set the learning rate to 5e-3 for the position and the scaling of 3D Gaussians and the remaining parameter learning rate is the same as vanilla 3D-GS \cite{kerbl20233d}. Particularly, the learning rate for the positions decays exponentially, reaching 0.01 times the initial value by the 60,000 iterations. From the 500 iterations until the 60000 iterations, we activate the densification strategy with binding inheritance every 100 iterations and reset the opacity every 3,000 iterations. 

After 5,000 iterations, we refine the Geometry Morph Adjuste using an Adam optimizer with $\beta=(0.9, 0.999)$ and add perceptual loss $\mathcal{L}_{\text{VGG}}$. The learning rates for both the basis network $\mathcal{F}_\mathcal{H}$ and the latent vector generation network $\mathcal{F}_{\psi, \theta}$ are set at 1e-4. The learning rate for the three-plane hash table is also set at 0.005.
\begin{figure*}
    \centering
    \includegraphics[width=1\linewidth]{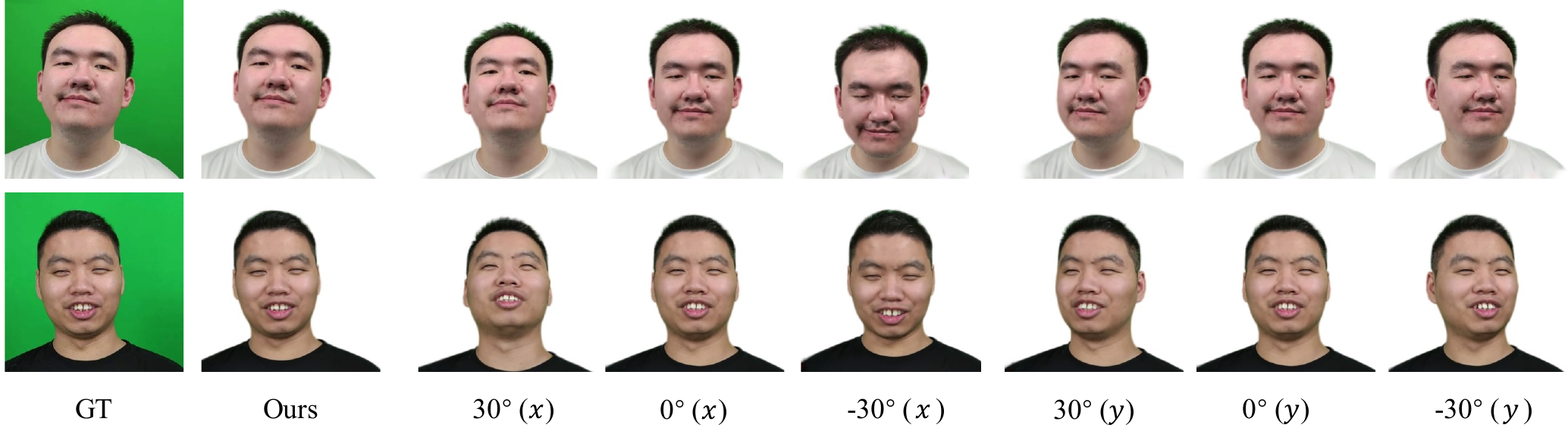}
    \caption{Novel view synthesis results of GGAvatar. We demonstrate multi-view geometric consistency across both x- and y-axis rotations.}
    \label{fig:4}
\end{figure*}
\begin{table*}[t] \centering
    \caption{Quantitative comparison with state-of-the-art methods. \colorbox{red!25}{  Red  } indicates the best and \colorbox{yellow!25}{ yellow } indicates the second.}
    \label{tab:table9}
    \resizebox{\textwidth}{!}{
    \Huge
        \begin{tabular}{c|*{3}{c}|*{3}{c}|*{3}{c}|*{3}{c}}
        \hline\hline
      \multicolumn{1}{c|}{Methods}  & \multicolumn{3}{c|}{PointAvatar} 
                               & \multicolumn{3}{c|}{ GaussianAvatars} 
                               & \multicolumn{3}{c|}{FlashAvatar} 
                               & \multicolumn{3}{c}{Ours}  \\
                               \hline
                              \multicolumn{1}{c|}{Metrics}  
                              & PSNR$\uparrow$ & 
                               SSIM $\uparrow$& 
                               LPIPS $\downarrow$
                               & PSNR$\uparrow$ & 
                               SSIM $\uparrow$& 
                               LPIPS $\downarrow$
                                & PSNR$\uparrow$ & 
                               SSIM $\uparrow$& 
                               LPIPS $\downarrow$
                               & PSNR$\uparrow$ & 
                               SSIM $\uparrow$& 
                               LPIPS $\downarrow$ \\
        \midrule
        ID 1    & 25.40 & 0.892 & 0.147
                & 27.63 & 0.918 & \cellcolor{yellow!25}0.089 
                & \cellcolor{yellow!25}32.00 & \cellcolor{yellow!25}0.938 & 0.121 
                & \cellcolor{red!25}33.61 & \cellcolor{red!25}0.956 & \cellcolor{red!25}0.060 \\
        ID 2    & 24.77 & 0.893 & 0.142
                & 20.29 & 0.899 & \cellcolor{yellow!25}0.106 
                & \cellcolor{yellow!25}26.67 & \cellcolor{yellow!25}0.922 & 0.123  
                & \cellcolor{red!25}28.23 &\cellcolor{red!25}0.939 & \cellcolor{red!25}0.079 \\
        ID 3    & 23.52 & 0.917 & 0.107
                & 24.31 & 0.936 & 0.088 
                & \cellcolor{red!25}24.99  & \cellcolor{yellow!25}0.938 & \cellcolor{yellow!25}0.087 
                & \cellcolor{yellow!25}24.76 & \cellcolor{red!25}0.942 & \cellcolor{red!25}0.064 \\
        ID 4    & 19.43 & 0.797 & 0.215
                & 19.51 & 0.836 & 0.186 
                & \cellcolor{yellow!25}25.58   & \cellcolor{yellow!25}0.917 &\cellcolor{yellow!25} 0.126 
                & \cellcolor{red!25}26.02 & \cellcolor{red!25}0.932 & \cellcolor{red!25}0.064 \\
        ID 5    & 23.76 & 0.933 & 0.070
                & \cellcolor{yellow!25}27.30 & \cellcolor{yellow!25}0.965 & \cellcolor{yellow!25}0.053 
                & 26.96   & 0.926 & 0.103 
                & \cellcolor{red!25}27.35 & \cellcolor{red!25}0.968 & \cellcolor{red!25}0.046 \\
        
        \hline\hline
    \end{tabular}
    }
\end{table*}
\section{Experiments}
\subsection{Datasets}
All our experiments are conducted on publicly available datasets released by previous work 
\cite{gao2022reconstructing} to ensure fair comparisons, with all data resized to a resolution of 512x512 in advance. The training data consists of approximately 2,500 to 3,000 frames. Furthermore, we select one frame every 20 frames from the original dataset to serve as test data, totaling around 200 frames. The binary mask was created using RVM \cite{lin2022robust} and employed for foreground segmentation.
\subsection{Baselines}
For comparison purposes, we select three state-of-the-art head avatar reconstruction methods: 1) PointAvatar \cite{zheng2023pointavatar}, which utilizes explicit point clouds with upsampling to construct head geometry; 2) GaussianAvatars \cite{qian2023gaussianavatars}, which binds Gaussians to the FLAME mesh, incorporating a mesh-driven approach to guide the deformation of Gaussian; and 3) Flashavatar \cite{xiang2023flashavatar}, which employs Gaussian initialization based on the UV plane and learns additional deformations to achieve high-fidelity digital human driving. For PointAvatar, we follow the author's suggestions to train it on a 32GB V100 GPU and use 240,000 points for all subject data. For a fair comparison, GaussianAvatars is trained under the same experimental conditions as ours, with an additional perceptual loss applied for supervision. Additionally, we set the UV resolution of FlashAvatar to 256 and added the boundary of the FLAME mesh.

\subsection{Qualitative and Quantitative Comparison in Reconstruction}
Figure~\ref{fig:3} shows the results of our visual comparison with the baseline rendering. PointAvatar reconstructs avatars using explicit point clouds and learns the LBS basis of each point to control point movement. However, this approach often fails to fit extreme expressions as illustrated in the first row. Furthermore, the primitives adopted are points with fixed shapes, inhibiting clear facial structure modeling. GaussianAvatars employs a standard LBS formula to control facial movements, which usually fails to model extreme expressions and dynamic high-frequency details. As evidence, observe the avatars generated by this method in the 1st and 4th rows, which demonstrate the actions of closing the eyes and the high-frequency details of the mouth, respectively.
FlashAvatar employs Gaussian primitives embedded at fixed positions within the FLAME mesh to initialize, which lacks adaptive control over the density of these primitives, resulting in visual artifacts such as visible cracks or seams, as seen in the mouth area in the first and second rows. This issue persists even when the head resolution is set to 256. Additionally, merely concatenating facial parameters and coordinates into an MLP is insufficient for capturing fine deformations, as demonstrated in the 3rd row (please refer to more renderings from it in the Ablation studies). Compared to the methods above, our rendering images more closely approximate the ground truth, achieving more accurate deformations and capturing nearly all dynamic facial details.

Quantitative result comparisons are recorded in Table~\ref{tab:table9}. We calculate metrics including PSNR, SSIM, and LPIPS \cite{Zhang_Isola_Efros_Shechtman_Wang_2018} separately for each subject. It is seen that our approach outperforms others by a significant margin.

\begin{figure}[h]
    \centering
    \includegraphics[width=1\linewidth]{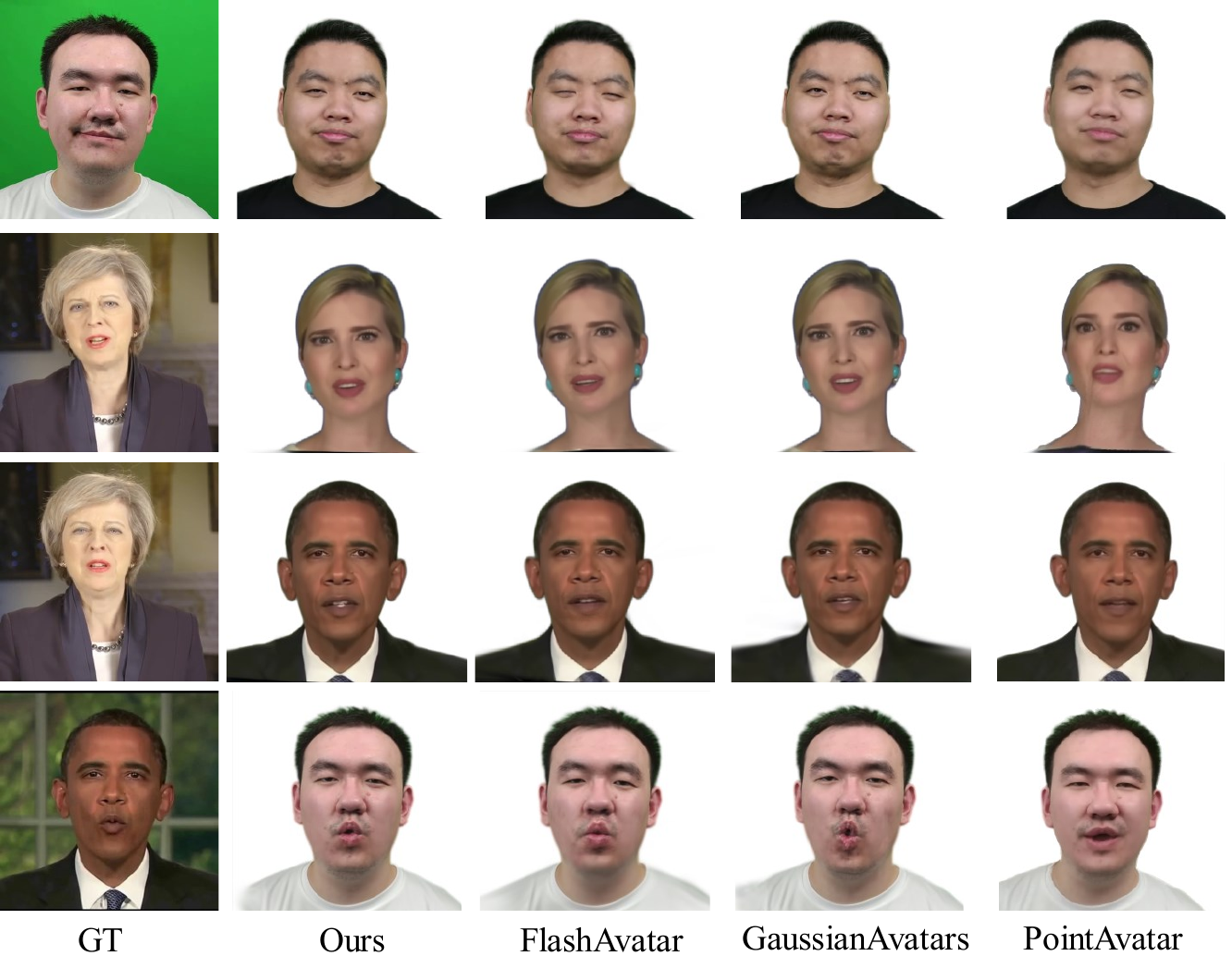}
    \caption{Cross-Identity Reenactment results comparison.}
    \label{fig:5}
\end{figure}

\subsection{Novel View Synthesis and Cross-Identity Reenactment}

To verify the robustness of the model, we designed a multi-view synthesis experiment for the GGAvatar model and a cross-identity reenactment experiment, comparing our results with the baseline methods. 

The results of the multi-view synthesis experiment are shown in Figure \ref{fig:4}. Our model can render images from different angles, producing images similar to the ground truth (GT). Our model accurately synthesized new perspectives among hundreds of pictures in the test set. This demonstrates the model's ability to maintain geometric consistency and high-fidelity detail reproduction across various viewing angles.

The comparison results with the baseline in cross-identity reenactment are shown in Figure \ref{fig:5}. Using Gaussian for expression parameter decoding, our model better replicates the original expressions onto new faces, capturing most details, including the corners of the eyes and mouth.

The baseline methods, such as FlashAvatar\cite{xiang2023flashavatar}, GaussianAvatars\cite{qian2023gaussianavatars}, and PointAvatar\cite{zheng2023pointavatar}, show limitations in accurately preserving these fine details. FlashAvatar tends to over-smooth facial features, losing critical expression nuances. GaussianAvatars, while preserving some facial contours, often distort subtle expression elements, leading to less natural results. PointAvatar, despite its geometric approach, struggles with maintaining expression fidelity across different identities.

Our model's ability to accurately transfer expressions while retaining fine details significantly enhances the realism of cross-identity reenactment. This superior performance is attributed to our advanced encoding and decoding algorithm, which precisely maps the expression parameters onto related 3d Gaussian structures. The results demonstrate the robustness and versatility of our approach, making it a promising solution for applications requiring high-fidelity facial reenactment.
\begin{table}[t] 
    \centering
    \caption{Ablation study on ID 1.  \colorbox{red!25}{  Red  } indicates the best and \colorbox{yellow!25}{ yellow } indicates the second.}
    \label{tab:2}
    \begin{tabular}{c|c|c|c}
        \hline\hline
        \multicolumn{1}{c|}{Methods}  
        & PSNR$\uparrow$ & SSIM $\uparrow$& LPIPS $\downarrow$ \\
        \midrule
        w/o Geometry Morph Adjuster & 28.43 & 0.926 & 0.092 \\
        w/o tri-plane & \cellcolor{yellow!25}33.41& 0.953 & \cellcolor{yellow!25} 0.061 \\
        w/o LBS & 33.01 & 0.952 & 0.062 \\
        w/o initialization & 33.34 & \cellcolor{yellow!25} 0.954 & 0.062 \\
        \hline
        Ours & \cellcolor{red!25}33.61 & \cellcolor{red!25}0.956 & \cellcolor{red!25}0.060 \\
        \hline\hline
    \end{tabular}
\end{table}
\subsection{Ablation Study}
To validate the effectiveness of our method components, we deactivate each of them and report results in Table~\ref{tab:2}.
\subsubsection*{Geometry Morph Adjuster}
Without the Geometry Morph Adjuster, the 3D avatar animations rely only on the linear transformations defined by the LBS formula. Although LBS-based deformations can robustly model coarse geometric changes, they fail to accurately capture high-frequency facial details, as shown in the 1st row of Table~\ref{tab:2}. It is worth noting that without the Geometry Morph Adjuster, our approach differs slightly from GaussianAvatars. To reduce training time, we do not employ the strategy of refining FLAME parameters throughout the training process of GGAvatar.
\subsubsection*{Parametric Tri-plane}
In the Geometry Morph Adjuster, we utilize a parameterized tri-plane hash table to store the geometric information of neutral Gaussians, where its multi-resolution grid encoding mechanism facilitates high-quality feature representation. For comparison, we replace the tri-plane with the positional encoding introduced by \cite{mildenhall2021nerf}, as shown in the 2nd row of Table~\ref{tab:2}.
\subsubsection*{Linear Blend Skinning}
Learning complex facial deformations from scratch is challenging. We use the Geometry Morph Adjuster to improve fitting efficiency to enhance the LBS-based deformations rather than directly learning the complete deformations. This allows our Geometry Morph Adjuster to focus only on high-frequency dynamic details. To demonstrate the effectiveness of our strategy, we performed an ablation study by removing the LBS and using the Geometry Morph Adjuster only to predict deformations. As shown in the 3rd row of Table~\ref{tab:2}, removing the LBS formula negatively impacts image quality and deformation results.
\subsubsection*{Neutral Gaussian Initialization}
We apply the Neutral Gaussian Initialization strategy before training our Geometry Morph Adjuster to model neutral Gaussian geometric shapes and accelerate convergence. To demonstrate the effectiveness of this strategy, we conducted an ablation study by removing the Neutral Gaussian Initialization. Specifically, we trained the Geometry Morph Adjuster from the beginning without the initialization phase of neutral Gaussians. As shown in the 4th row of Table~\ref{tab:2}, without the Neutral Gaussian Initialization phase, the Geometry Morph Adjuster struggles to capture fine-grained deformation details. Therefore, an effective Neutral Gaussian Initialization strategy is crucial for achieving higher deformation accuracy and fidelity.


\section{Conclusion and Discussion}
In this paper, we introduce GGAvatar, a coarse-to-fine framework that integrates the strengths of LBS formulation and deformable basis learning. Our approach consists of two core modules: the Neutral Gaussian Initialization module and the Geometry Morph Adjuster. The Neutral Gaussian Initialization module quickly enhances avatar details using Gaussian primitives, while the Geometry Morph Adjuster employs a parameterized multi-resolution tri-plane and a finely tuned MLP to adaptively predict and optimize modeling and deformation areas that linear LBS cannot accurately control. It is worth noting that our method, to some extent, overcomes the limitations of FLAME, using only its topological deformation (LBS) as a prior for the deformation process.

Extensive testing on public datasets demonstrates that GGAvatar outperforms existing methods in both visual quality and quantitative metrics. By maintaining high fidelity in dynamic head movements and complex expressions, our method ensures realistic and expressive avatar reenactment.

\section{Limitations and Future Work}
Despite the promising results, our model has several limitations. First, the reliance on the FLAME model means that the accuracy of FLAME parameter tracking is crucial. In this paper, we did not improve the tracker, resulting in some entanglement between identity and facial parameters. 

In some extreme facial expression scenarios, the Gaussians transformed by LBS may cluster together, leading to a significant overlap of Gaussian features. This overlap can result in unexpected outcomes that cannot be adequately adjusted even with our Geometry Morph Adjuster. Additionally, the Gaussian distribution of the scene may become abnormal. In future work, we hope to develop a better adaptive method to handle these extreme scenarios and intricate details, aiming to achieve a more robust 3D digital human reconstruction method.

\bibliographystyle{eg-alpha-doi} 
\bibliography{EGauthorGuidelines-PG2024-fin}       


\newpage


\end{document}